\documentclass{article}
\usepackage{spconf,amsmath,graphicx}
\usepackage{graphicx}
\usepackage{multicol}
\usepackage{multirow}
\usepackage{amsmath}
\usepackage{enumitem}
\usepackage[numbers,sort]{natbib}

\title{DLM-VMTL:A double layer mapper for Heterogeneous Data Video Multi-task Prompt Learning}
\name{Zeyi Bo, Wuxi Sun, Ye Jin}
\address{Harbin Institute of Technology}

\begin{document}
%
\maketitle
\begin{abstract}
In recent years, the parameters of backbones of Video Understanding tasks continue to increase and even reach billion-level. Whether fine-tuning a specific task on the Video Foundation Model or pre-training the model designed for the specific task, incurs a lot of overhead. How to make these models play other values than their own tasks becomes a worthy question. Multi-Task Learning(MTL) makes the visual task acquire the rich shareable knowledge from other tasks while joint training. It is fully explored in Image Recognition tasks especially dense predict tasks. Nevertheless, it is rarely used in video domain due to the lack of multi-labels video data. In this paper, a heterogenous data video multi-task prompt learning (VMTL) method is proposed to address above problem. It's different from it in image domain, a Double-Layers Mapper(DLM) is proposed to extract the shareable knowledge into visual promptS and align it with representation of primary task. Extensive experiments prove that our DLM-VMTL performs better than baselines on 6 different video understanding tasks and 11 datasets.

\end{abstract}
\begin{keywords}
Video Understanding, Multi-Task Learning, Prompt Learning, Heterogenous Data
\end{keywords}
\section{Introduction}
\label{sec:intro}

Recent years, the Transformer architecture which drives the development of Computer Vision models\cite{3,4,5,6,7} is dominant. There are many variants of Vision Transformer(VIT) as ASformer\cite{7}, MTFormer\cite{3}, MulT\cite{4}, STAR\cite{6} as the backbones of different Video Understanding tasks. However, as parameters of the model continue to expand, a problem is arising: Full fine-tuning a Video Transformer on the specific downstream task incured significant overhead,  How can the excellent capabilities of the model be transferred for other tasks' models to benefit them? 
\begin{figure}
    \centering
    \includegraphics[width=0.48\textwidth, height=4cm]{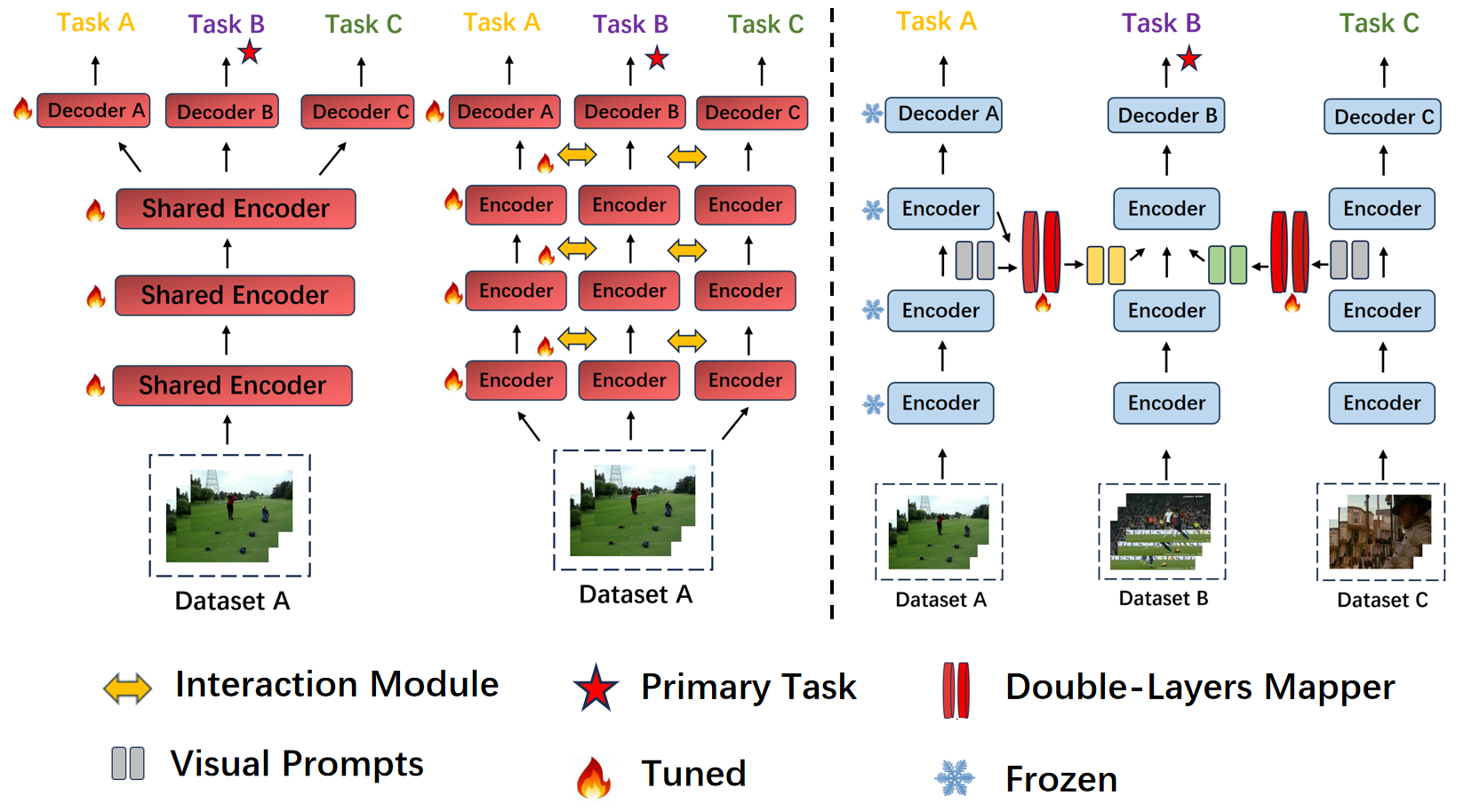}
    \caption{The pipelines of conventional Multi-Task Learning\textbf{(left)} proposed new paradigm for Heterogeneous Data Video Multi-Task Prompt Learning\textbf{(right)}.
    }
    \label{fig1}
\end{figure}

Multi-task Learning(MTL)\cite{17} jointly train many visual tasks to make each task acquire the shareable knowledge while minimizing the conflict caused by the joint training. The conventional MTL can be categorized into two types, the Hard Parameters Sharing(HPS) and the Soft Parameters Sharing(SPS). To conclude, the summarised pipeline of conventional MTL is shown in left in Fig \ref{fig1}. The HPS\cite{3,4,18} makes a shared backbone network as a feature extractor for all tasks which supervised by the loss function of all tasks. Despite HPS can improve the deployment efficiency to meet some project needs, this leads to the gradient conflict and performance degradation for some tasks. The SPS\cite{5,19,20} makes each task use its own pipeline and utilizes a variety of interaction modules for information interaction to get shareable knowledge between tasks during training. Although the performance improved, SPS involves a large number of parameters during training and inferring and requires a large number of memory for deployment. Moreover, whether HPS or SPS has another problem: the inputs of models must be the same just different tasks correspond to different labels. In image domain, especially dense prediction tasks, there are datasets\cite{22,23} customized for MTL. However, in video domain, the cost of annotations is much higher than that of image. The multi-labels dataset is difficult to obtain, so that the conventional MTL methods are not suitable for the Video Understanding tasks. 

Nowadays, there is a novel way of MTL, the Multi-Task Learning across datasets. AdaMV-MOE\cite{9} copied the MLP layer as the experts group and used the extended model and the learnable routing algorithm for Multi-Task Learning across datasets. However, this method enlarged the model several times, and the routing algorithm was difficult to train and converge. \cite{10} introduces shareable knowledge by transferring visual prompts trained on auxiliary task groups for primary task as non-random initializations of prompts. However, this method requires manual grouping of similar tasks and it is invalid for some task groups. And there was no study of the Video Understanding tasks. 

VPT \cite{11} first proposed the visual prompt learning which insert the learnable prompts into the representation of embedding layer. However, Insert the prompts too soon maybe has a negative impact on some primary tasks\cite{10} and prompts trained from auxiliary task groups is used directly to initialize the primary task’s prompts can lead to missing representation alignment between heterogeneous tasks. \cite{12} prove that the representation from intermediate layer of VIT contain a wealth of task-specific features that are not decoded by subsequent layers. Therefore, prompts learned from the intermediate layer instead of patch embedding layer can learn a wealth of task-specific visual information. \cite{13,14} prove that Heterogeneous representation need some kind of mappings to align.

In this paper, we propose a new paradigm for Heterogeneous Data Video Multi-Task Prompt Learning(VMTL). A Double-Layer Mapper(DLM-VMTL) is proposed to learn the knowledgeable prompts from the pre-trained video models on auxiliary tasks for primary task with frozen backbones. The pipeline as shown in right in Fig \ref{fig1}. The first layer utilizes the self-attention mechanism to calculate the prompts which interaction with intermediate layers’ representation of the auxiliary task, and get the beneficial prompts for primary task. The second layer maps the prompts to align with the representations of primary task. The prompts from DLM-VMTL can be used to prompt the middle layers’ representation of the primary task to acquire the shareable knowledge provided by the auxiliary tasks.

In summary, our contribution are as follows:

\begin{itemize}
  \item We propose a new paradigm for Heterogeneous Data Video Multi-Task Prompt Learning. It can improve the performance and generalization of the primary task with 10.8$\%$ of total parameters. Some pre-trained task-specific Video Understanding models are successfully utilized for other tasks.
  \item We propose the Double-Layers Mapper(DLM-VMTL), a plug and play module which can learn the prompts from the intermediate layers' representation of the auxiliary tasks to provide the shareable knowledge to primary task. It is directly supervised by the loss function of primary task so it has target consistency with the primary task.
  \item After extensive experiments, our DLM-VMTL outperform the baselines on 11 datasets from 6 Video Understanding tasks. To the best of our knowledge, it is the first time to combine the prompt learning with Video MTL.
\end{itemize}

\section{Related Work}
\subsection{Multi-Task Learning}
\label{ssec:subhead}

The conventional MTL can be categorized into HPS and SPS. They are fully explored in image domain but are not applicable in video domain because of the deficient multi-labels data. Multi-Task Learning across datasets is a promising solution. \cite{10} first proposed the use of Prompt learning to transfer one model's capabilities pre-trained with task A to another model whose primary task is B. However, it is ineffective for some tasks because of its oversimplified pre-training way for prompts. And there is no study of the Video Understanding tasks.

\subsection{Prompt Tuning}
\label{ssec:subhead}

In recent years, Prompt Tuning is popular in Computer Vision. \cite{15} used learnable text Prompts to prompt the CLIP\cite{25} model for better inference accuracy. \cite{11} proposed a purely visual prompts to fine-tune the frozen model for the first time and achieve competitive results on image recognition tasks. \cite{16} abstracts adjacent frames into purely visual prompts and interactively computes them with the representation of the current frame to introduce timing feature to the model. \cite{10} used Prompts pre-trained on the auxiliary tasks to fine-tune the primary task to realize MTL. However, it is ineffective for some tasks. \cite{13,14} states that some kind of mappings are needed to align representation across tasks and models. Method in \cite{10} lacks the re-abstraction of the transferred prompts.

\begin{figure*}[t]
\includegraphics[width=0.7\textwidth]{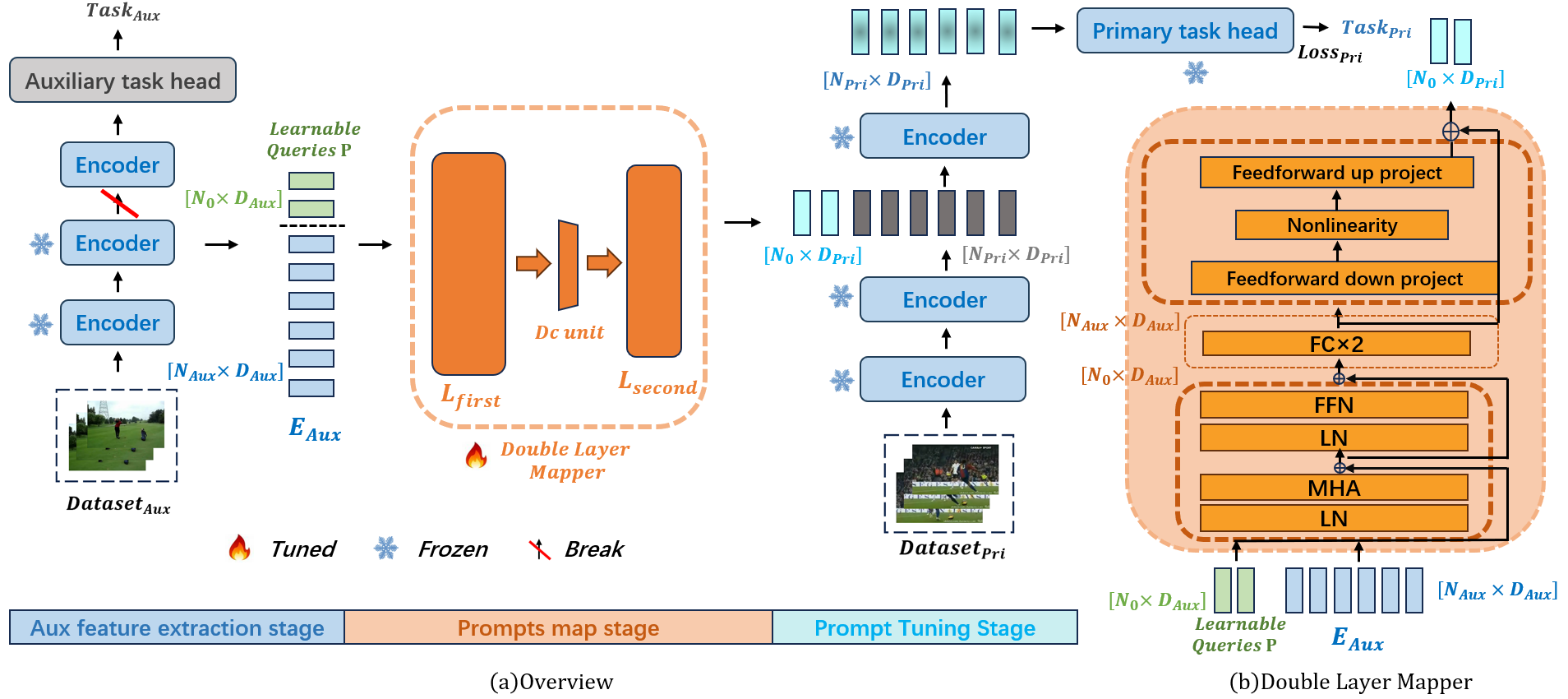}
\centering
\caption{(a) An overview of our DLM-VMTL. Remarkably, we describe case where one auxiliary task prompts the primary task. The case of multiple auxiliary tasks adds more prompts at prompt tuning stage. (b) The detailed structure of DLM.}
\label{fig2}
\end{figure*}

\section{Methods}
The pipeline for our proposed DLM-VMTL is shown in Fig \ref{fig2}. We use one auxiliary task to prompt the primary task as example. The same can be said for more auxiliary tasks, more auxiliary tasks introduce more prompts at prompt tuning stage to prompt the primary task. The first layer DLM-VMTL will be introduced in Section \ref{sub3.1}. The Dimension corresponding unit will be introduced in Section \ref{sub3.2}. The second layer of DLM-VMTL will be introduced in Section \ref{sub3.3}.

\subsection{The first layer of DLM-VMTL}
\label{sub3.1}
This part of DLM-VMTL plays a role in learning cross-task knowledge and transfering the knowledge to prompts. It is a Transformer-based structure which proved effective for prompts learning in \cite{16}. The input prompts $P$$\in$$R^{N_{0}\times D_{Aux}}$ are the learnable parameters. Cross-task knowledge prompts $P_{Aux}^{i}$$\in$$R^{N_{0}\times D_{Aux}}$ are generated through the self-attention calculation with middle layer representation of auxiliary task as Eq \ref{eq1}.$N_0$ denotes the number of tokens of prompts. $L$ consists of Multihead Self-Attention (MSA) and Feed-Forward Networks (FFN) together with LayerNorm and residual connections\cite{8}. $N$ denotes the number of the auxiliary tasks. Suppose $n_i$ represents the number of layers of the encoder for $task_i$, $middle$ = $n_i$/2. 

\begin{equation}
    [P_{Aux}^i,E_{middle+1}]=L([P,E_{middle}]) \quad
    i = 1,2,...,N
    \label{eq1}
\end{equation}

\subsection{The Dimension corresponding unit} 
\label{sub3.2}
When $P_{Aux}^{i}$ is generated, the Dimension corresponding unit matches its dimension with the dimension of representation of primary task. It consists of two fully connected layers instead of one, this design inspired by \cite{39} that reduces the number of parameters.As Eq \ref{eq2}, the structure of $FC_1$ is $D_{Aux}$$\times$$d_0$, the structure of $FC_1$ is $d_0$$\times$$D_{Pri}$ Let $d_0$ = 16, the parameters can be reduced by about 95$\%$.Finally, the deminsion of $P_{Aux}^{i}$ changes from $N_0$$\times$$D_{Aux}$ to $N_0$$\times$$D_{Pri}$.

\begin{equation}
    P_{Aux}^i = FC_2(FC_1(P_{Aux}^i))
    \label{eq2}
\end{equation}

\subsection{The second layer of DLM-VMTL}
\label{sub3.3}
Due to the heterogeneity of the auxiliary tasks and primary task models, some mapping alignment is required between $P_{Aux}^{i}$ and primary task representation before calculation. Here we use an Adapter\cite{39} module from the NLP field for alignment operations. Since our DLM-VMTL is not used in the inference phase, we can take advantage of its efficient computation and avoid its disadvantage of increasing the inference overhead. As Eq \ref{eq3} and Eq \ref{eq4}, we align $P_{Aux}^{i}$ with the primary task to generate $P_{pri}^{i}$ and connect $P_{pri}^{i}$ from all auxiliary tasks as $P_{pri}$. As Eq \ref{eq5}, the $P_{pri}$ prompts the primary task and $L_{pri}$ denotes the Encoder layer of primary task. Our target is to train the excellent $P_{pri}^{i}$ so that it extracts cross-task knowledge aligned with the primary task. That is, to train our DML-VMTL which is directly supervised by the loss of primary task, so that it has target consistency with the primary task.

\begin{equation}
    P_{Pri}^i = Adapter(P_{Aux}^i)
    \label{eq3}
\end{equation}

\begin{equation}
    P_{Pri}=concat(P_{Pri}^1,P_{Pri}^2,...,P_{Pri}^N) \quad
    i = 1,2,...,N
    \label{eq4}
\end{equation}

\begin{equation}
    [P,E_{middle+1}^{Pri}]=L_{Pri}([P_{Pri},E_{middle}^{Pri}])
    \label{eq5}
\end{equation}

\begin{table*}[t]
\small
\begin{center}
\caption{Comparison with aforementioned baselines. Without additional description, the evaluation metric is MAP.} \label{tab:cap}
\begin{tabular}{|l|c|c|c|c|c|c|r}
\hline
  \multirow{2}*{Method}& \multicolumn{6}{|c|}{Primary task} \\
\cline{2-7}
         & SSv2(TOP 1) & AVA & Thumos14& 50salads(F1$@$50$\%$) & ImageNet VID & YouTube-VIS 2019 \\
\hline     
    Full-finetuned(100$\%$) & 75.2 & 40.4 &67.8& 76.0 & 90.1 &64.9 \\   
\hline
    VPT-Shallow(0.21$\%$) & 74.1 & 38.9 &66.1& 74.3 & 88.1 &62.7 \\   
\hline
    HPS(33.3$\%$) & 71.3 & 36.9 &64.4& - & - &- \\   
\hline
    MVLPT(0.21$\%$) & 74.7 & 39.5 &66.4& 73.9 & 88.3 &62.3 \\   
\hline
    DLM-VMTL(10.8$\%$) & \textbf{76.7} & \textbf{42.2} &\textbf{68.6} & \textbf{79.5} & \textbf{91.0} &\textbf{66.1}\\   
\hline
\end{tabular}
\label{tab1}
\vspace{-2.0em}
\end{center}
\end{table*}

\begin{table*}[t]
\small
\begin{center}
\caption{Comparision with the full-tinetuned method on Zero-shot transfer ability.} \label{tab:cap}
\begin{tabular}{|l|c|c|c|c|c|r}
\hline
  \multirow{2}*{}& \multicolumn{5}{|c|}{Zero-shot transfer} \\
\cline{2-6}
         & UCF101(TOP 1) & HMDB51(TOP 1) & AVA-kinetics& Breakfast(F1$@$50$\%$) & Gtea(F1$@$50$\%$)  \\
\hline     
    Original & 97.7 & 85.1 &41.9& 53.7 & 75.8  \\   
\hline
    DLM-VMTL & \textbf{98.6} & \textbf{86.9} &\textbf{43.0}& \textbf{56.5} & \textbf{78.3}  \\   
\hline

\end{tabular}
\label{tab2}
\vspace{-2.0em}
\end{center}
\end{table*}

\section{Experiments}

We will introduce the video task used in the experiments, the corresponding datasets and the backbones. The Baselines compared to our DLM-VMTL and the details of the experiment implementation will also be presented. Section \ref{sub4.1} describes the effectiveness of our DLM-VMTL on each dataset, Section  \ref{sub4.2} describes the excellent generalization ability of the model with the prompts trained by our DLM-VMTL, and Section  \ref{sub4.3} describes the ablation experiments of our DLM-VMTL.

\textbf{Dataset} We tested our mehotd with 11 different datasets on six different video tasks.
Video action recognition: We use HMDB51\cite{26}, UCF101\cite{27}, SSv2\cite{28}. The primary task is SSv2, the UCF101 and HMDB51 to verify the model generalization ability.Spatial action detection: We used AVA\cite{29}, AVA-kinetics\cite{30}, The primary task is AVA and the AVA-kinetics verify the generalization of the model .Temporal Action Detection: We use Thumos14 \cite{31}.Action Segmentation: We used 50salads\cite{32}, Breakfast\cite{33} and gtea\cite{34}.  Breakfast and gtea verify the generalization of the model.Video Object Detectoin(VOD): We use ImageNet-VID\cite{35}. Video Instance Segmentation(VIS):We use YouTube-VIS 2019\cite{36}

\textbf{Baseline}  1) The full-finetuned backbone of each task.The backbone of Video Action recognition and Spatial action detection and Temporal Action Detection is the VIT-B which distilled by pre-trained VIT-g\cite{2}, following the pipeline in \cite{2}. The Action Segmentation uses ASFormer\cite{7}, the VOD uses TransVOD\cite{37}. The VIS uses DVIS\cite{38}. 2) The single-task prompt learning. Use the VPT\cite{11} for the pre-trained backbone of each task to prompt tuning. 3) The simple HPS for 3 tasks on VIT-B. 4) Multi-task Prompt learning in \cite{10} for each task, the MVLPT.

\textbf{Implementation Details} For HPS, the Video Action Recognition and Spatial Action Detection and Temporal Action Detection are jointly trained because they have the same backbone. For VPT, we only use VPT-shallow beacuse of it consistents with our method. For MVLPT, we follow the way in \cite{10}, only use few-shot training set from source task and we select the task after the primary task in Table 1 as the source task. During training, if one task is primary task, the other five tasks are auxiliary tasks.
\subsection{DLM Video Multi-Task Prompt Learning}
\label{sub4.1}
We experiment on all the 6 tasks to verify the performance of the proposed DLM-VMTL for Prompt Learning . The results are summarized in Table \ref{tab1}.It can be seen that the simple HPS has a serious impact on the VIT-B in multi-task learning across datasets because of the way of across datasets. MVLPT\cite{10} shows more inefficiency on video tasks which are more complex than image tasks. Our DLM-VMTL Prompt Learning works for all primary tasks. Compared with full-finetuned, our DLM-VMTL uses only 10.8$\%$ of the parameters of all backbones but achieves superior performance on all 6 video tasks.

\begin{figure}
    \centering
    \includegraphics[width=0.48\textwidth, height=3.6cm]{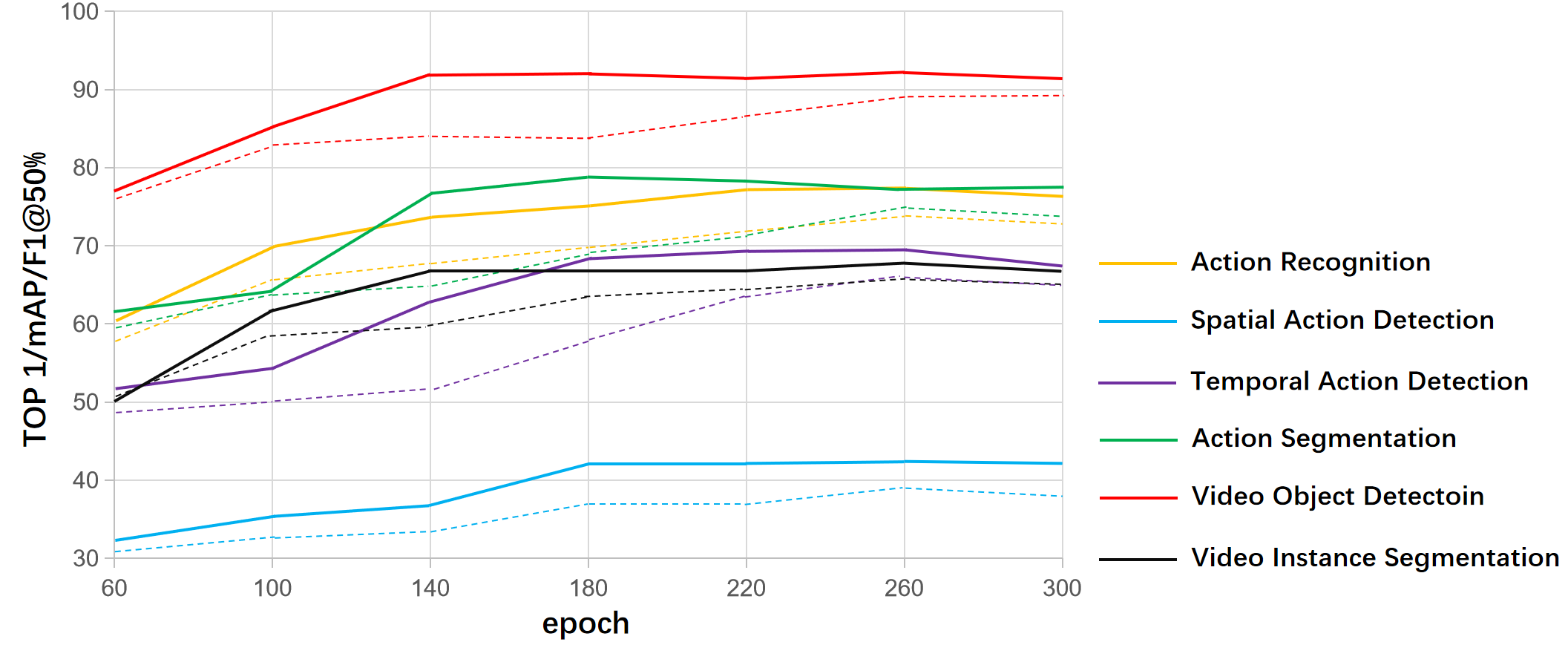}
    \caption{Comparison between single-layer and double-layer structures. Dashed lines denotes single-layer structure.
    }
    \label{fig3}
\end{figure}

\subsection{Zero-shot transfer performance}
\label{sub4.2}
The prompts trained by DMP-VMTL can not only improve the performance of the primary task, but also improve the Zero-shot transfer capability. We save prompts trained by auxiliary tasks for some primary tasks and load them in Zero-shot tansfer test on other datasets related to each primary task. The results are compared with the fine-tuned model without prompts. As the results shown in Table \ref{tab2}, the model is more knowledgeable and no longer ignorant of the unknown domain after our DLM Video Multi-Task Prompt Learning.
\subsection{Ablation Study }
\label{sub4.3}
We conducted ablation experiment on the effectiveness of the second layer of DLM. The result is shown in Fig \ref{fig3}. The solid line represents the use of the original DLM and the dashed line represents the removal of the second layer from the original DLM. It can be seen that the Double-Layer structure which aligns prompts with the representation of primary task not only results in better final performance, but also speeds up convergence.

\section{Conclusions}
In this paper, we propose a new paradigm for Heterogeneous Data Video Multi-Task Prompt Learning.  It utilizes the proposed Double-Layer Mapper to learn cross-task knowledge for improving the performance on 11 datasets with a little parameters. We hope our study will inspire future research on video Multi-Task Learning.

\bibliographystyle{ieeetr}

\bibliography{1,2,3,4,5,6,7,8,9,10,11,12,13,14,15,16,17,18,19,20,22,23,24,25,26,27,28,29,30,31,32,33,34,35,36,37,38,39}

\end{document}